\documentclass{article} 
\usepackage{iclr2022_conference,times}

\usepackage{amsmath,amsfonts,bm}

\def\eqref#1{equation~\ref{#1}}

\def\1{\bm{1}}

\DeclareMathAlphabet{\mathsfit}{\encodingdefault}{\sfdefault}{m}{sl}
\SetMathAlphabet{\mathsfit}{bold}{\encodingdefault}{\sfdefault}{bx}{n}

\usepackage{hyperref}

\usepackage{url}
\usepackage{algorithm}
\usepackage{algorithmic}

\usepackage{microtype}
\usepackage{graphicx}
\usepackage{subfigure}
\usepackage{booktabs} 
\usepackage{multirow}
\usepackage{tabularx}
\usepackage{tabu}

\usepackage{amsmath}
\usepackage{amssymb}
\usepackage{mathtools}
\usepackage{amsthm}
\usepackage{multirow}
\usepackage{arydshln}

\renewcommand{\algorithmicrequire}{ \textbf{Input:}} 
\renewcommand{\algorithmicensure}{ \textbf{Output:}} 

\title{Modeling Adversarial Noise for Adversarial Training}

\iclrfinalcopy

\author{
Dawei Zhou$^{* 1}$,
Nannan Wang\thanks{Equal contribution.}\text{ } $^{1}$, 
Bo Han$^{2}$,
Tongliang Liu\thanks{Corresponding author.}\text{ } $^{3}$ \\
$^1$Xidian University;
$^2$Hong Kong Baptist University;
$^3$The University of Sydney\\
}

\begin{document}

\maketitle

\begin{abstract}
Deep neural networks have been demonstrated to be vulnerable to adversarial
noise, promoting the development of defense against adversarial attacks. Motivated by the fact that adversarial noise contains well-generalizing features and that the relationship between adversarial data and natural data can help infer natural data and make reliable predictions, in this paper, we study to model adversarial noise by learning the transition relationship between adversarial labels (i.e. the flipped labels used to generate adversarial data) and natural labels (i.e. the ground truth labels of the natural data). Specifically, we introduce an instance-dependent transition matrix to relate adversarial labels and natural labels, which can be seamlessly embedded with the target model (enabling us to model stronger adaptive adversarial noise). Empirical evaluations demonstrate that our method could effectively improve adversarial accuracy.
\end{abstract}

\section{Introduction}
\label{section1}
Deep neural networks have been demonstrated to be vulnerable to adversarial noise \cite{goodfellow2014explaining,szegedy2013intriguing,shen2017ape,liao2018defense,ma2018characterizing,wu2020adversarial}. The vulnerability of deep neural networks seriously threatens many decision-critical deep learning applications \cite{lecun1998gradient,he2016deep,Zagoruyko2016WRN,simonyan2014very,2017Mask,ma2021understanding}.

To alleviate the negative affects caused by adversarial noise, many adversarial defense methods have been proposed. A major class of adversarial defense methods focus on exploiting adversarial instances to help train the target model \cite{madry2017towards,ding2019sensitivity,zhang2019theoretically,wang2019improving}, which achieve the state-of-the-art performance. However, these methods do not explicitly model the adversarial noise. The relationship between the adversarial data and natural data has not been well studied yet.

Studying (or modeling) the relationship between adversarial data and natural data is considered to be beneficial. If we can model the relationship, we can infer natural data information by exploiting the adversarial data and the relationship. Previous researches have made some explorations on this idea. Some data pre-processing based methods \cite{shen2017ape,liao2018defense,naseer2020self,pmlr-v139-zhou21e, zhou2021removing}, which try to recover the natural data by removing the adversarial noise, share the same philosophy. However, those methods suffer from the high dimensionality problem because both the adversarial data and natural data are high dimensional. The recovered data is likely to have \textit{human-observable loss} (i.e., obvious inconsistency between processed instances and natural instances) \cite{xu2017feature} or contain residual adversarial noise \cite{liao2018defense}. 

To avoid the above problems, in this paper, we propose to model adversarial noise in the low-dimensional label space. Specifically, instead of directly modeling the relationship between the adversarial data and the natural data, we model the adversarial noise by learning the label transition from the \textit{adversarial labels} (i.e., the flipped labels used to generate adversarial data) to the \textit{natural labels} (i.e., the ground truth labels of the natural data). Note that the adversarial labels have not been well exploited in the community of adversarial learning, which guide the generation of adversarial noise and thus contain valuable information for modeling the well-generalizing features of the adversarial noise \cite{ilyas2019adversarial}. 

It is well-known that the adversarial noise depends on many factors, e.g., the data distribution, the adversarial attack strategy, and the target model. The proposed adversarial noise modeling method is capable to take into account of the factors because that the label transition is dependent of the adversarial instance (enabling us to model how the patterns of data distribution and adversarial noise affect label flipping) and that the transition can be seamlessly embedded with the target model (enabling us to model stronger adaptive adversarial noise). Specifically, we employ a deep neural network to learn the complex instance-dependent label transition. 

By using the transition relationship, we propose a defense method based on Modeling Adversarial Noise (called MAN). Specifically, we embed a label \textit{transition network} (i.e., a matrix-valued transition function parameterized by a deep neural network) into the target model, which denotes the transition relationship from mixture labels (mixture of adversarial labels and natural labels) to natural labels (i.e., the ground truth labels of natural instances and adversarial instances). The transition matrix explicitly models adversarial noise and help us to infer natural labels.
Considering that adversarial data can be adaptively generated, we conduct joint adversarial training on the target model and the transition network to achieve an optimal adversarial accuracy. We empirically show that the proposed MAN-based defense method could provide significant gains in the classification accuracy against adversarial attacks in comparison to state-of-the-art defense methods.  

The rest of this paper is organized as follows. In Section~\ref{section2}, we introduce some preliminary information and briefly review related work on attacks and defenses. In Section~\ref{section3}, we discuss how to design an adversarial defense by modeling adversarial noise. Experimental results are provided in Section~\ref{section4}. Finally, we conclude this paper in Section~\ref{section5}.

\section{Preliminaries}
\label{section2}

In this section, we first introduce some preliminary about notation and the problem setting. We then review the most relevant literature on adversarial attacks and adversarial defenses.

\noindent\textbf{Notation.} 
We use \textit{capital} letters such as $X$ and $Y$ to represent random variables, and \textit{lower-case} letters such as $x$ and $y$ to represent realizations of random variables $X$ and $Y$, respectively. For norms, we denote by $\|x\|$ a generic norm. Specific examples of norms include $\|x\|_{\infty}$, the $L_{\infty}$-norm of $x$, and $\|x\|_{2}$, the $L_2$-norm of $x$. Let $\mathbb{B}(x, \epsilon)$ represent the neighborhood of $x$: $\{\tilde{x}:\|\tilde{x}-x\| \leq \epsilon$\}, where $\epsilon$ is the perturbation budget. We define the \textit{classification function} as $f: \mathcal{X} \rightarrow \{1,2,\ldots,C\}$, where $\mathcal{X}$ is the feature space of $X$. It can be parametrized, e.g., by deep neural networks.

\noindent\textbf{Problem setting.} 
This paper focuses on a classification task under the adversarial environment, where adversarial attacks are utilized to craft adversarial noise to mislead the prediction of the target classification model. Let $X$ and $Y$ be the variables for natural instances and natural labels (i.e., the ground truth labels of natural instances) respectively. We sample natural data $\{(x_i, y_i)\}_{i=1}^{n}$ according to the distribution of the variables $(X,Y)$, where $(X,Y) \in \mathcal{X} \times\{1,2, \ldots, C\}$. Given a deep learning model based classifier $f$ and a pair of natural data $(x,y)$, the adversarial instance $\tilde{x}$ satisfies the following constraint:
\begin{equation}
\label{eq1}
f\left(\tilde{x}\right) \neq y \quad \text { s.t.} \quad\left\|x-\tilde{x}\right\| \leq \epsilon \text{.}
\end{equation}
The generation of the adversarial instance is guided by the adversarial label $\tilde{y}$ which is different with the natural label $y$. Adversarial labels can be set by the attacker in target attacks, or be randomly initialized and then found by non-target attacks. Let $\widetilde{X}$ and $\widetilde{Y}$ be the variables for adversarial instances and adversarial labels respectively. We denote by $\{(\tilde{x}_i, \tilde{y}_i)\}_{i=1}^{n}$ the adversarial data drawn according to a distribution of the variables $(\widetilde{X},\widetilde{Y})$. Our aim is to design an adversarial defense to correct the adversarial label $\tilde{y}$ into natural label $y$ by modeling the relationship between adversarial data (i.e., $\tilde{x}$ or $\tilde{y}$) and natural data (i.e., $x$ or $y$).

\noindent\textbf{Adversarial attacks.} 
Adversarial instances are inputs maliciously designed by adversarial attacks to mislead deep neural networks \cite{szegedy2013intriguing}. They are generated by adding imperceptible but adversarial noise to natural instances. Adversarial noise can be crafted by multi-step attacks following the direction of adversarial gradients, such as PGD \cite{madry2017towards} and AA \cite{croce2020reliable}. The adversarial noise is bounded by a small norm-ball $\|\cdot\|_{p} \leq \epsilon$, so that their adversarial instances can be perceptually similar to natural instances. Optimization-based attacks such as CW \cite{carlini2017towards} and DDN \cite{rony2019decoupling} jointly minimize the perturbation $L_2$ and a differentiable loss based on the logit output of a classifier. Some attacks such as FWA \cite{wu2020stronger} and STA \cite{xiao2018spatially} focus on mimicking non-suspicious vandalism by exploiting the geometry and spatial information. 

\noindent\textbf{Adversarial defenses.} The issue of adversarial instances promotes the development of adversarial defenses. A major class of adversarial defense methods is devoted to enhance the adversarial robustness in an adversarial training manner \cite{madry2017towards,ding2019sensitivity,zhang2019theoretically,wang2019improving}. They augment training data with adversarial instances and use a min-max formulation to train the target model \cite{madry2017towards}. However, these methods do not explicitly model the adversarial noise. The relationship between the adversarial data and natural data has not been well studied yet.

In addition, some data pre-processing based methods try to remove adversarial noise by learning a denoising function or a feature-squeezing function. For example, denoising based defenses \cite{liao2018defense,naseer2020self} transfer adversarial instances into clean instances, and feature squeezing based defenses \cite{guo2017countering} aim to reduce redundant but adversarial information. However, these methods suffer from the high dimensionality problem because both the adversarial data and natural data are high dimensional. For examples, the recovered instances are likely to have significant inconsistency between the natural instances \cite{xu2017feature}. Besides, the recovered instances may contain residual adversarial noise \cite{liao2018defense}, which would be amplified in the high-level layers of the target model and mislead the final predictions. To avoid the above problems, we propose to model adversarial noise in the low-dimensional label space. 

\section{Modeling adversarial noise based defense}
\label{section3}
In this section, we presen the \textit{Modeling Adversarial Noise} (MAN) based defense method, to improve the adversarial accuracy against adversarial attacks. We first illustrate the motivation of the proposed defense method (Section~\ref{section3.1}). Next, we introduce how to model adversarial noise (Section~\ref{section3.2}). Finally, we present the training process of the proposed adversarial defense (Section~\ref{section3.3}). The code is available at \url{https://github.com/dwDavidxd/MAN}.

\subsection{Motivation}
\label{section3.1}
Studying the relationship between adversarial data and natural data is considered to be beneficial for adversarial defense. If we can model the relationship, we can infer clean data information by exploiting the adversarial data and the relationship. Processing the input data to remove the adversarial noise is a representative strategy to estimate the relationship. However, data pre-processing based methods may suffer from high dimensionality problems. To avoid the problem, we would like to model adversarial noise in the low-dimensional label space.

Adversarial noise can be modeled because it have imperceptible but well-generalizing features. Specifically, classifiers are usually trained to solely maximize accuracy for natural data. They tend to use any available signal including those that are well-generalizing, yet brittle. Adversarial noise can arise as a result of perturbing these features \cite{ilyas2019adversarial}, and it controls the flip from natural labels to adversarial labels. Thus, adversarial noise contains well-generalizing features which can be modeled by \textit{learning the label transition from adversarial labels to natural labels}.

Note that the adversarial noise depends on many factors, such as data distribution, the adversarial attack strategy, and the target model. Modeling adversarial noise in label space is capable to take into account of these factors. Specifically, since that the label transition is dependent of the adversarial instance, we can model how the patterns of data distribution and adversarial noise affect label flipping, and exploit the modeling to correct the flipping of the adversarial label. In addition, we can model adversarial noise crafted by the stronger white-box adaptive attack, because the label transition can be seamlessly embedded with the target model. By jointly training the transition matrix with the target model, it can also be adaptive to adversarial attacks.
We employ a deep neural network to learn the complex label transition from the well-generalizing and misleadingly predictive features of instance-dependent adversarial noise.

Motivated by the value of modeling adversarial noise for adversarial defense, we propose a defense method based on Modeling Adversarial Noise (MAN) by exploiting label transition relationship. 

\subsection{Modeling adversarial noise}
\label{section3.2}
We exploit a transition network to learn the label transition for modeling adversarial noise. The transition network can be regarded as a matrix-valued transition function parameterized by a deep neural network. The transition matrix, estimated by the label transition network, explicitly models adversarial noise and help us to infer natural labels from adversarial labels. The details are discussed below.

\begin{figure}[t]
   \vskip 0.1in
    \centering
    \subfigure[]{
        \label{fig1_1}
        \includegraphics[width=1.7in]{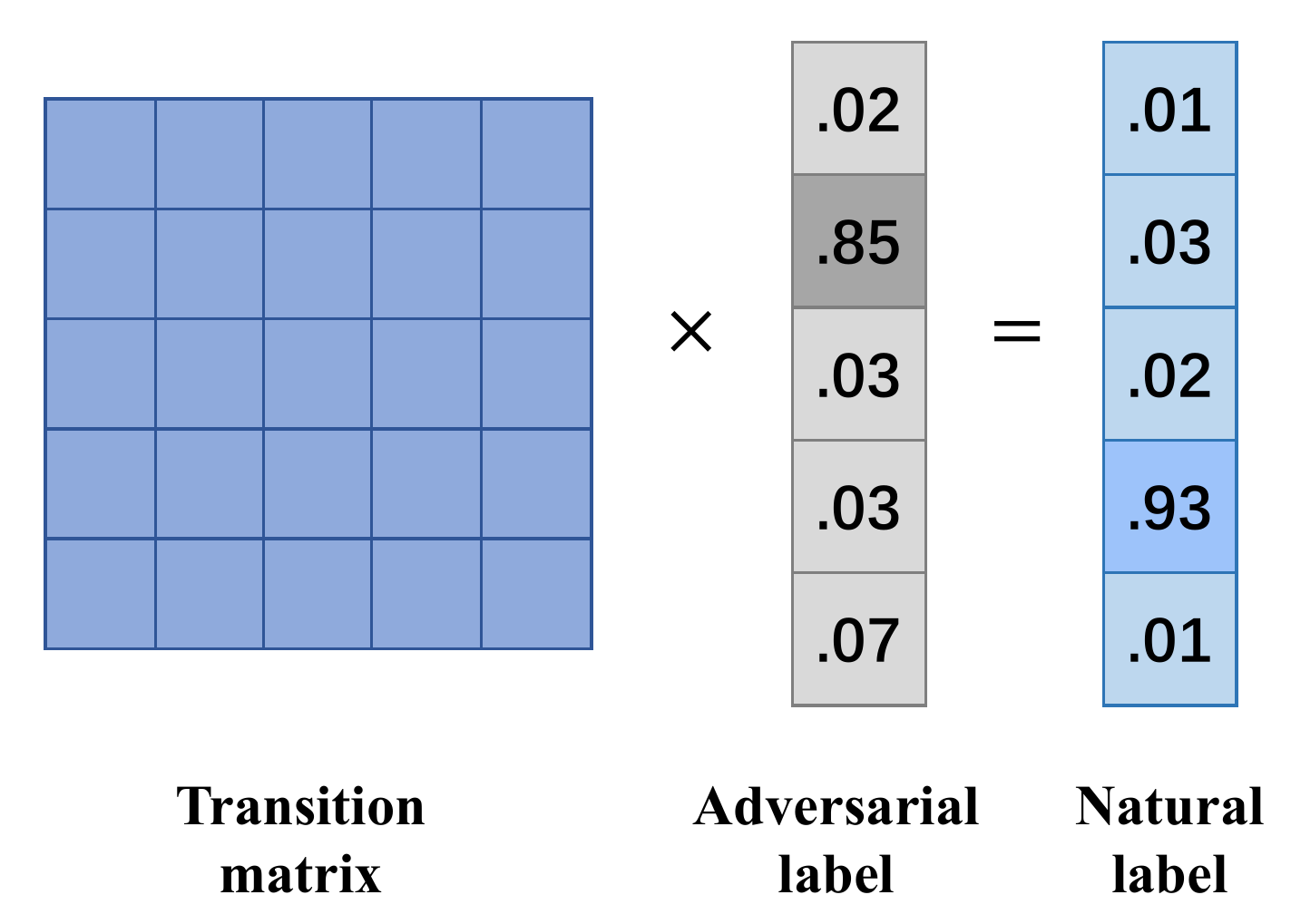}}
    \hspace{10mm}
    \subfigure[]{
        \label{fig1_2}
        \includegraphics[width=2.7in]{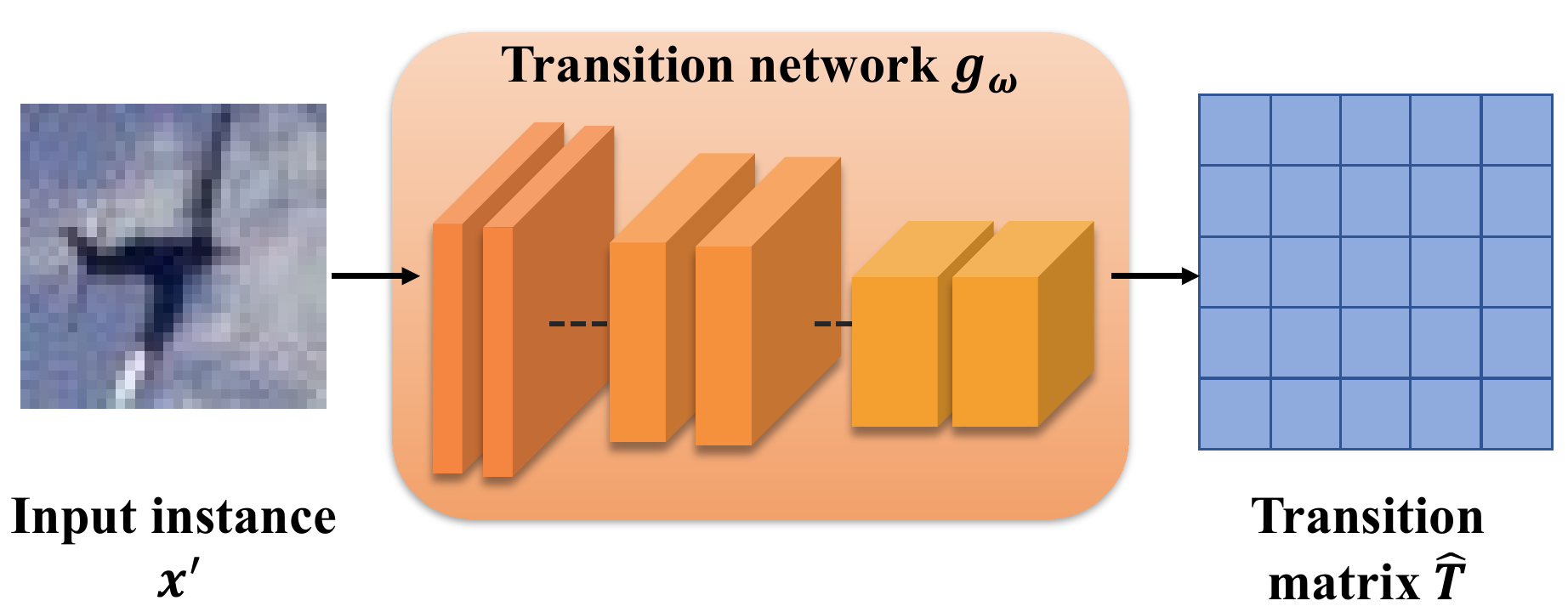}}
    \caption{(a). Infer the natural label by utilizing the transition matrix and the adversarial label. (b). The transition network $g_{\omega}$ parameterized by $\omega$ takes the instance $x^{\prime}$ as input, and output the label transition matrix $\widehat{T}(x^{\prime};\omega)=g_{\omega}(x^{\prime})$.}
\end{figure}

\begin{figure*}[t]
    \vskip 0.1in
    \centering    
    \includegraphics[width=5.7in]{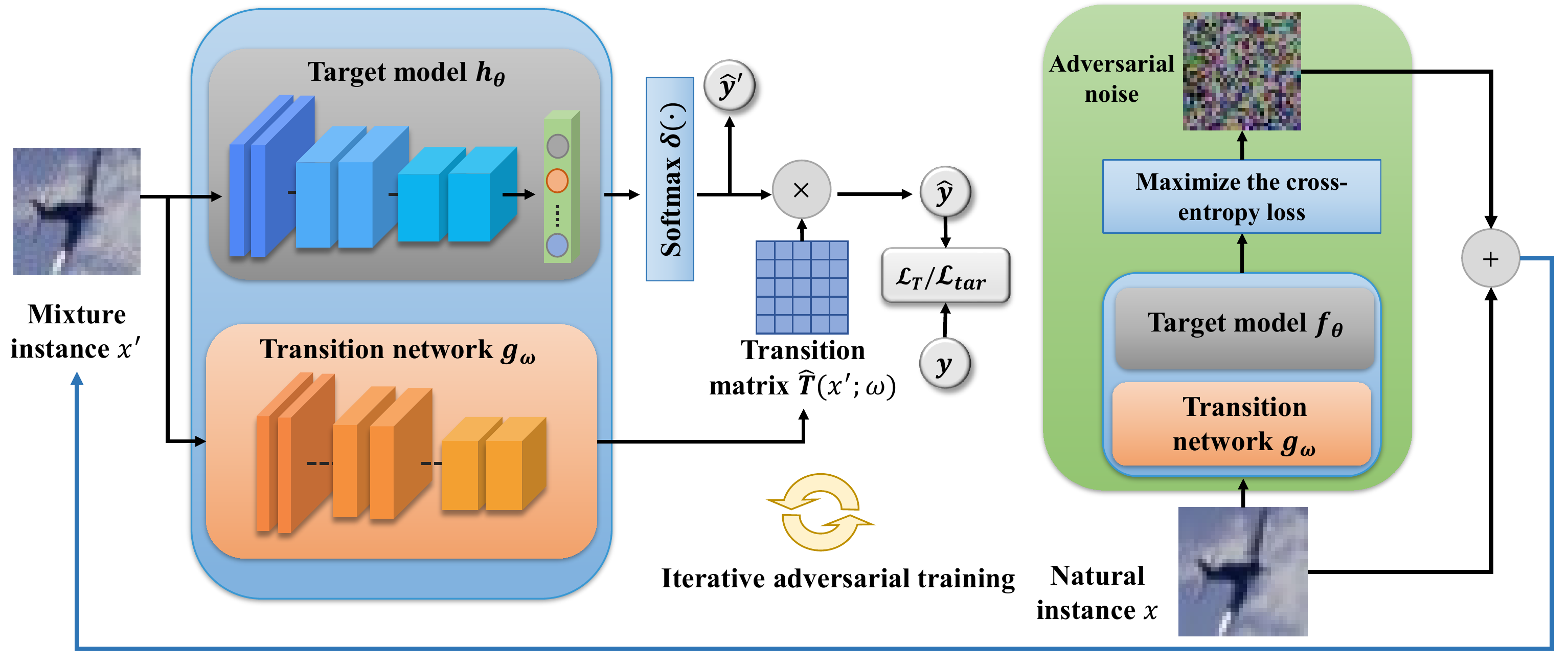}
    \caption{An overview of the training procedure for the proposed defense method. $\boldsymbol{\hat{y}^{\prime}}$ and $\boldsymbol{\hat{y}}$ denote the probability of the estimated mixture label $\hat{y}^{\prime}$ and the probability of the inferred natural label $\hat{y}$ respectively, i.e., $\boldsymbol{\hat{y}^{\prime}}=\delta(h_{\theta}(x^{\prime}))$ and $\boldsymbol{\hat{y}}=\boldsymbol{\hat{y}^{\prime}} \cdot \widehat{T}(x^{\prime};\omega)$. $\boldsymbol{y}$ is $y$ in the form of a vector.}
    \label{fig2}
\end{figure*}

\noindent\textbf{Transition matrix.} 
To model the adversarial noise, we need to relate adversarial labels and natural labels in a explicit form (e.g., a matrix). Inspired by recent researches in label-noise learning \cite{xia2020part,liu2015classification,xia2019anchor,yang2021estimating,wu2021class2simi,xia2021robust}, we design a label transition matrix, which can encode the probabilities that adversarial labels flip into natural labels. We then can infer natural labels by utilizing the transition matrix and adversarial data (see Fig.~\ref{fig1_1}). Note that the transition matrix used in our work is different from that used in label-noise learning. The detailed illustration can be found in Appendix~\ref{appendix_1}.

Considering that the adversarial noise is instance-dependent, the transition matrix should be designed to depend on input instances, that is, we should find a transition matrix specific to each input instance. In addition, we not only need to focus on the accuracy of adversarial instances, but also the accuracy of natural instances. An adversarially robust target model is often trained with both natural data and adversarial data. We therefore exploit the mixture of natural and adversarial instances (called mixture instances) as input instances, and design the transition matrix to model the relationship between the mixture of natural and adversarial labels (called mixture labels) to the ground-truth labels of mixture instances (called natural labels).

Specifically, let $X^{\prime}$ denote the variable for the mixture instances, $Y^{\prime}$ denote the variable for the mixture labels, and $Y$ denote the variable for the natural labels. We combine the natural data and adversarial data as the mixture data, i.e.,  $\{(x_i^{\prime}, y_i^{\prime})\}_{i=1}^{2n}=\{(x_i, y_i)\}_{i=1}^{n} \cup \{(\tilde{x_i}, \tilde{y_i})\}_{i=1}^{n}$, where $\{(x^{\prime}_i, y^{\prime}_i)\}_{i=1}^{2n}$ is the data drawn according to a distribution of the random variables $(X^{\prime},Y^{\prime})$. Since we have the ground-truth labels of mixture instances (i.e., natural labels), we can extend the mixture data $\{(x_i^{\prime}, y_i^{\prime})\}_{i=1}^{2n}$ into a triplet form $\{(x^{\prime}_i, y^{\prime}_i,y_i)\}_{i=1}^{2n}$. We utilize the transition matrix $T \in[0,1]^{C\times C}$ to model the relationship between mixture labels $Y^{\prime}$ and natural labels $Y$. $T$ is dependent of the mixture instances $X^{\prime}$. It is defined as:
\begin{equation}
\label{eq2}
T_{i, j}(X^{\prime}=x^{\prime})=P\left(Y=j \mid Y^{\prime}=i, X^{\prime}=x^{\prime}\right) \text{,}
\end{equation}
where $T_{i, j}$ denotes the $(i,j)-th$ element of the matrix $T(X^{\prime}=x^{\prime})$. It indicates the probability of the mixture label $i$ flipped to the natural label $j$ for the input $x^{\prime}$. 

The basic idea is that given the \textit{mixture class posterior probability} (i.e., the class posterior probability of the mixture labels $Y^{\prime}$) $P(\boldsymbol{Y}^{\prime} \mid X^{\prime}=x^{\prime})=[P(Y^{\prime}=1 \mid X^{\prime}=x^{\prime}), \ldots, P(Y^{\prime}=C \mid X^{\prime}=x^{\prime})]^{\top}$, the \textit{natural class posterior probability} (i.e., the class posterior probability of natural labels $Y$) $P(\boldsymbol{Y} \mid X^{\prime}=x^{\prime})$ could be inferred by exploiting $P(\boldsymbol{Y}^{\prime} \mid X^{\prime}=x^{\prime})$ and $T(X^{\prime}=x^{\prime})$:
\begin{equation}
\label{eq3}
P(\boldsymbol{Y} \mid X^{\prime}=x^{\prime})=T(X^{\prime}=x^{\prime})^{\top}P(\boldsymbol{Y}^{\prime} \mid X^{\prime}=x^{\prime}) \text{.}
\end{equation}
We then can obtain the natural labels by choose the class labels that maximize the robust class posterior probabilities. Note that the mixture class posterior probability $P(\boldsymbol{Y}^{\prime} \mid X^{\prime}=x^{\prime})$ can be estimated by exploiting the mixture data.

\noindent\textbf{Transition network.}  
we employ a deep neural network (called \textit{transition network}) to estimate the label transition matrix by exploiting the mixture data $\{(x^{\prime}_i, y^{\prime}_i,y_i)\}_{i=1}^{2n}$. Specifically, as shown in Fig.~\ref{fig1_2}, the transition network $g_{\omega}(\cdot)$ parameterized by $\omega$ takes the mixture instance $x^{\prime}$ as input, and output the label transition matrix $\widehat{T}(x^{\prime};\omega)=g_{\omega}(x^{\prime})$. We then can infer the natural class posterior probability $P(\boldsymbol{Y} \mid X^{\prime}=x^{\prime})$ according to Eq.~\ref{eq3}, and thus infer the natural label. To optimize the parameter $\omega$ of the transition network, we minimize the difference between the inferred natural labels and the ground-truth natural labels. The loss function of the transition network is define as:

\begin{equation}
\label{eq4}
\mathcal{L}_{T}(\omega)=-\frac{1}{2n} \sum_{i=1}^{2n} \ell (\boldsymbol{y^{\prime}_i} \cdot \widehat{T}(x^{\prime}_{i};\omega), \boldsymbol{y_{i}}) \text{,}
\end{equation}
where $\boldsymbol{y^{\prime}_i}$, $\boldsymbol{y_{i}}$ are $y^{\prime}_i$, $y_{i}$ in the form of vectors. $\ell(\cdot)$ is the cross-entropy loss between the inferred natural labels and the ground-truth natural labels, i.e., $\ell(\boldsymbol{y^{\prime}_i} \cdot \widehat{T}(x^{\prime}_{i};\omega), \boldsymbol{y_{i}})=  \boldsymbol{y_i} \cdot \log(\boldsymbol{y^{\prime}_i} \cdot \widehat{T}(x_i^{\prime};\omega))$. 

\subsection{Training}
\label{section3.3}
Considering that adversarial data can be adaptively generated, and the adversarial labels are also depended on the target model, we conduct joint adversarial training on the target model and the transition network to achieve the optimal adversarial accuracy. We provide the overview of the training procedure in Fig.~\ref{fig2}.

For the transition network, we optimize its model parameter $\omega$ according to Eq.~\ref{eq4}. For the target model, considering that the final inferred natural class posterior probability is influenced by the target model, we also use the cross-entropy loss between the inferred natural labels and the ground-truth natural labels to optimize the parameter $\theta$. The loss function for the target model $h_{\theta}$ is defined as:
\begin{equation}
\label{eq5}
\begin{gathered}
\mathcal{L}_{tar}(\theta)=-\frac{1}{2n} \sum_{i=1}^{2n} [\boldsymbol{y_i} \cdot \log(\boldsymbol{\hat{y}^{\prime}_i} \cdot \widehat{T}(x_i^{\prime};\omega))] \text{,} \\ 
\boldsymbol{\hat{y}^{\prime}_i}=\delta(h_{\theta}(x^{\prime}_{i})) \text{,} 
\end{gathered}
\end{equation}
where $\delta(\cdot)$ denote the softmax function. $\boldsymbol{\hat{y}^{\prime}_i}$ denote the mixture label in the form of a vector predicted by the target model.

The details of the overall procedure are presented in Alg.~\ref{alg1}. Specifically, for each mini-batch $\mathcal{B}=\{x_i\}_{i=1}^{m}$ sampled from natural training set, we first generate adversarial instances $\widetilde{\mathcal{B}}=\{\tilde{x}_i\}_{i=1}^{m}$ via a strong adversarial attack algorithm, and obtain the mixture mini-batch $\mathcal{B}^{\prime}=\{x_i^{\prime}\}_{i=1}^{2m}$ (i.e., mixture of $\mathcal{B}$ and $\widetilde{\mathcal{B}}$) with mixture labels $\{y_{i}^{\prime}\}_{i=1}^{2m}$ (i.e., mixture of adversarial labels $\{\tilde{y}_i\}_{i=1}^{m}$ and natural labels $\{y_i\}_{i=1}^{m}$). Then, we input the mixture instances $\{x_i^{\prime}\}_{i=1}^{2m}$ into the target model $h_{\theta}(\cdot)$ and output $\{\boldsymbol{\hat{y}_{i}^{\prime}}\}_{i=1}^{2m}$. We also input the mixture instances $\{x_i^{\prime}\}_{i=1}^{2m}$ into the transition network $g_{\omega}(\cdot)$ to estimate the label transition matrices $\{\widehat{T}(x_i^{\prime};\omega)\}_{i=1}^{2m}$. We next infer the final prediction labels by exploiting $\{\boldsymbol{\hat{y}_{i}^{\prime}}\}_{i=1}^{2m}$ and $\{\widehat{T}(x_i^{\prime};\omega)\}_{i=1}^{2m}$. Finally, we compute the loss functions $\mathcal{L}_T(\omega)$ and $\mathcal{L}_{tar}(\theta)$, and update the parameters $\omega$ and $\theta$. By iteratively conducting the procedures of adversarial instance generation and defense training, $\omega$ and $\theta$ are expected to be adversarially optimized.

To demonstrate the effectiveness of joint adversarial training, we conduct an ablation study by independently training the transition network with a fixed pre-trained target model. In addition, to prove that the improvement of our method is not mainly due to the introduction of more model parameters, we conduct an additional experiment by using an adversarially trained target model fused by two backbone networks as the baseline. The details could be found in Section~\ref{section4.3}.

\begin{algorithm}[t]
   \caption{\small Training the defense model based on Modeling Adversarial Noise (MAN).}
   \label{alg1}
\begin{algorithmic}[1]
   \REQUIRE Target model $h_{\theta}(\cdot)$ parameterized by $\theta$, transition network $g_{\omega}(\cdot)$ parameterized by $\omega$, batch size $m$, and the perturbation budget $\epsilon$;
   \REPEAT
   \STATE Read mini-batch $\mathcal{B}=\{x_i\}_{i=1}^{m}$ from training set;
   \STATE Craft adversarial instance $\{\tilde{x}_i\}_{i=1}^{m}$ at the given perturbation budget $\epsilon$ for each instance $x_i$ in $\mathcal{B}$;
   \STATE Obtain the mixture mini-batch $\mathcal{B}^{\prime}=\{x_i^{\prime}\}_{i=1}^{2m}$ with mixture labels $\{y_{i}^{\prime}\}_{i=1}^{2m}$;
   \FOR{$i=1$ to $2m$ (in parallel)}
   \STATE Forward-pass $x_i^{\prime}$ through $h_{\theta}(\cdot)$ and obtain $\boldsymbol{\hat{y}_{i}^{\prime}}$;
   \STATE Forward-pass $x_i^{\prime}$ through $g_{\omega}(\cdot)$ to estimate $\widehat{T}(x_{i}^{\prime};\omega)$;
   \STATE Infer the final prediction label by exploiting $\boldsymbol{\hat{y}_{i}^{\prime}}$ and $\widehat{T}(x_i^{\prime};\omega)$;
   \ENDFOR
   \STATE Calculate $\mathcal{L}_T(\omega)$ and $\mathcal{L}_{tar}(\theta)$ using Eq.~\ref{eq4} and Eq.~\ref{eq5};
   \STATE Back-pass and update $\omega$ and $\theta$; 
   \UNTIL training converged.
\end{algorithmic}
\end{algorithm}

\begin{table*}[hbtp]
\caption{Adversarial accuracy (percentage) of defense methods against white-box adaptive attacks on \textit{CIFAR-10} and \textit{Tiny-ImageNet}. The target model is ResNet-18. We show the most successful defense with \textbf{bold}.}
\label{tab1}
\renewcommand\tabcolsep{7pt}
\renewcommand\arraystretch{1.0}
\begin{center}
\begin{tabular}{l|l|cccccc}
\hline
Dataset &Defense & None & PGD-40 & AA & FWA-40 & CW$_2$ & DDN \\ \hline
\multirow{6}{*}{CIFAR-10} &AT & \textbf{83.39} & 42.38 & 39.01 & 15.44 &0.00 & 0.09 \\
&MAN & 82.72 & \textbf{44.83} & \textbf{39.43} & \textbf{29.53} &\textbf{43.17} & \textbf{10.63} \\ \cdashline{2-8}[3pt/5pt]
&TRADES & \textbf{80.70} & 46.29 & 42.71 & 20.54 &0.00 & 0.06 \\
&MAN\_TRADES &80.34  &\textbf{48.65}  &\textbf{44.40}  &\textbf{29.13} &\textbf{1.46} &\textbf{0.31} \\ \cdashline{2-8}[3pt/5pt]
&MART & \textbf{78.21} & 50.23 & 43.96 & 25.56 &0.02 & 0.07  \\
&MAN\_MART &77.83 &\textbf{50.95}  &\textbf{44.42}  &\textbf{31.23} &\textbf{1.53} &\textbf{0.47} \\ \hline
\multirow{6}{*}{Tiny-ImageNet} &AT & \textbf{48.40} & 17.35 & 11.27 & 10.29 &0.00 & 0.29 \\
&MAN & 48.29 & \textbf{18.15} & \textbf{12.45} & \textbf{13.17} &\textbf{16.27} & \textbf{4.01} \\ \cdashline{2-8}[3pt/5pt]
&TRADES & \textbf{48.25} & 19.17 & 12.63 & 10.67 &0.00 & 0.05 \\
&MAN\_TRADES &48.19  &\textbf{20.12}  &\textbf{12.86}  &\textbf{14.91} &\textbf{0.67} &\textbf{1.10} \\ \cdashline{2-8}[3pt/5pt]
&MART & \textbf{47.83} & 20.90 & 15.57 & 12.95 &0.00 & 0.06  \\
&MAN\_MART &47.79  &\textbf{21.22}  &\textbf{15.84}  &\textbf{15.10} &\textbf{0.89} &\textbf{1.23} \\ \hline
\end{tabular}
\end{center}
\vskip -0.2in
\end{table*}

\section{Experiments}
\label{section4}
In this section, we first introduce the experiment setup in Section~\ref{section4.1}. Then, we evaluate the effectiveness of our defense against representative and commonly used $L_{\infty}$ norm and $L_2$-norm adversarial attacks inSection~\ref{section4.2}. In addition, we conduct ablation studies in Section~\ref{section4.3}. Finally, we present that MAN is also suitable for detecting adversarial samples Section~\ref{section4.4}

\subsection{Experiment setup}
\label{section4.1}
\noindent\textbf{Datasets.}
We verify the effective of our defense method on two popular benchmark datasets, i.e., \textit{CIFAR-10} \cite{krizhevsky2009learning} and \textit{Tiny-ImageNet} \cite{wu2017tiny}. \textit{CIFAR-10} has 10 classes of images including 50,000 training images and 10,000 test images. \textit{Tiny-ImageNet} has 200 classes of images including 100,000 training images, 10,000 validation images and 10,000 test images. Images in the two datasets are all regarded as natural instances. All images are normalized into [0,1], and are performed simple data augmentations in the training process, including random crop and random horizontal flip. For the target model, we mainly use ResNet-18 \cite{he2016deep} for both \textit{CIFAR-10} and \textit{Tiny-ImageNet}. 

\noindent\textbf{Attack settings.}
Adversarial samples for evaluating defense models are crafted by applying state-of-the-art attacks. These attacks include $L_{\infty}$-norm PGD \cite{madry2017towards}, $L_{\infty}$-norm AA \cite{croce2020reliable}, $L_{\infty}$-norm FWA \cite{wu2020stronger}, $L_2$-norm CW \cite{carlini2017towards} and $L_2$-norm DDN \cite{rony2019decoupling}. Among them, the AA attack algorithm integrates three non-target attacks and a target attack. Other attack algorithms belong to non-target attacks. The iteration number of PGD and FWA is set to 40 with step size 0.007. The iteration numbers of CW$_2$ and DDN are set to 200 and 40 respectively with step size 0.01. For \textit{CIFAR=10} and \textit{Tiny-ImageNet}, the perturbation budgets for $L_2$-norm attacks and $L_{\infty}$-norm attacks are $\epsilon=0.5$ and $8/255$ respectively.  

\noindent\textbf{Defense settings.}
We use three representative defense methods as the baselines: standard adversarial training method AT \cite{madry2017towards}, optimized adversarial training methods TRADES \cite{zhang2019theoretically} and MART \cite{wang2019improving}. For all baselines and our defense method, we use the $L_{\infty}$-norm non-target PGD-10 (i.e., PGD with iteration number of 10) with random start and step size $\epsilon/4$ to craft adversarial training data. The perturbation budget $\epsilon$ is set to $8/255$ for both \textit{CIFAR-10} and \textit{Tiny-ImageNet}. All the defense models are trained using SGD with momentum 0.9 and an initial learning rate of 0.1. For our defense method, we exploit the ResNet-18 as the transition network for both \textit{CIFAR-10} and \textit{Tiny-ImageNet}. Other detailed settings can be found in Appendix~\ref{appendix_2}.


\subsection{Defense effectiveness}
\label{section4.2}
\textbf{Defending against adaptive attacks.}
A powerful \textit{adaptive attack} strategy has been proposed to break defense methods \cite{athalye2018obfuscated,carlini2017magnet}. In this case, the attacker can access the architecture and model parameters of both the target model and the defense model, and then can design specific attack algorithms. We study the following three adaptive attack scenarios for evaluating our defense method.

\textit{\textbf{Scenario (i): disturb the final output.}} Considering that the models in baselines (i.e., the target models) are completely leaked to the attacker in the white-box setting, for fair comparison, we utilize white-box adversarial attacks against the combination of the target model and the transition matrix. Similar to attacks against baselines, the goal of the non-target attack in this scenario is to maximize the distances between final predictions of our defense and the ground-truth natural labels. The adversarial instance $\tilde{x}$ is crafted by solving the following optimization problem:
\begin{equation}
\label{eq6}
\begin{gathered}
\max _{\tilde{x}} \, \mathcal{L}(\boldsymbol{\tilde{y}} \cdot \widehat{T}(\tilde{x};\omega), \boldsymbol{y}) \text{,} \\ \text{ subject to: }\left\|x-\tilde{x}\right\| \leq \epsilon \text{,}
\end{gathered}
\end{equation}
where $\boldsymbol{\tilde{y}}=\delta(h_{\theta}(\tilde{x}))$ and $\mathcal{L}(\cdot)$ denote the specific loss function used by each attack. Similarly, we can generate adversarial instances via the target attack. The details of the target attack are presented in Appendix~\ref{appendix_3_1}.

We combine this attack strategy with five representative adversarial attacks introduced in Section \textit{Attack settings} to evaluate defenses. The average natural accuracy (i.e., the results in the third column) and the average adversarial accuracy of defenses are shown in Tab.~\ref{tab1}. 

The results show that our defense (i.e., MAN) achieves superior adversarial accuracy compared with AT. This presents that our defense is effective. Although our method has a slight drop in the natural accuracy (0.80\%), it provides more gains for adversarial robustness (e.g., 5.78\% against PGD-40 and 1.08\% against stronger AA). In addition, our method achieves significant improvements against some more destructive attacks (e.g., the adversarial accuracy is increased from 15.44\% to 29.53\% against FWA-40 and from 0.09\% to 10.63\% against DDN). The standard deviation is shown in Appendix~\ref{appendix_3_1}. Besides, we evaluate the effectiveness of our defense method on other model architecture by using the VggNet-19 as the target model and the transition network. In addtion, we evaluate the robustness performance of our defense method at a small batch-size (e.g., 128). These detailed results are also shown in Appendix~\ref{appendix_3_1}. 

Note that the training mechanism in our method can be regarded as the standard adversarial training on the combination of the target network and the transition matrix, our method thus is applicable to different adversarial training methods. To avoid the bias caused by different adversarial training methods, we apply the optimized adversarial training methods TRADES and MART to our method respectively (i.e., MAN\_TRADES and MAN\_MART). As shown in Tab.~\ref{tab1}, the results show that our method can improve the adversarial accuracy. Although our method has a slight drop in the natural accuracy (e.g., 0.49\% on \textit{CIFAR-10} and 0.08\% on \textit{Tiny-ImageNet} for MART), it provides more gains for adversarial robustness (e.g., 1.43\% and 1.53\% against PGD-40 on \textit{CIFAR-10} and \textit{Tiny-ImageNet} respectively). Besides, we find that using TRADES and MART affects the improvement of defense effectiveness against $L_2$-norm based attacks (e.g., CW$_2$ and DDN). We will further study and address this issue in the future work.

\textit{\textbf{Scenario (ii): attack the transition matrix.}} In this scenario, we design an adversarial attack to destroy the crucial transition matrix of our defense method. If the transition matrix is destroyed, the defense would immediately become ineffective. Since the ground-truth transition matrix is not given, we use the target attack strategy to craft adversarial samples. We choose an anti-diagonal identity matrix (see Fig.~\ref{fig5} in Appendix~\ref{appendix_3_2}) as an example of the target transition matrix in the target attack. The optimization goal is designed as:
\begin{equation}
\label{eq7}
\begin{gathered}
\max _{\tilde{x}} \, -\mathcal{L}_{mse}(\widehat{T}(\tilde{x};\omega), T^{*}) \text{,} \\ \text{ subject to: }\left\|x-\tilde{x}\right\| \leq \epsilon \text{,}
\end{gathered}
\end{equation}
where $\mathcal{L}_{mse}$ denote the \textit{mean square error} loss. 

We use $L_{\infty}$-norm PGD-40 with this target attack strategy to evaluate the MAN-based defense trained in scenario (i). The adversarial accuracy is 70.18\% on \textit{CIFAR-10}, which represents that our defense is effective against such type of adaptive target attack. This may be because that the attack in scenario (i) also tries to craft adversarial data by destroying the transition matrix to reduce the adversarial accuracy. The transition network adversarially trained with such adversarial data thus have good robustness against the attack designed for the transition matrix.

\textit{\textbf{Scenario (iii): dual attack.}} We explore another  adaptive attack scenario. In this scenario, the attack not only disturbs the final prediction labels, but also disturbs the output of the target model. We call such attack \textit{dual adaptive attack}. The optimization goal of the non-target dual attack can be designed as:
\begin{equation}
\label{eq8}
\begin{gathered}
\max _{\tilde{x}} \, [\mathcal{L}(\boldsymbol{\tilde{y}} \cdot T(\tilde{x};\omega), \boldsymbol{y}) + \mathcal{L}_{ce}(\boldsymbol{\tilde{y}}, \boldsymbol{y})] \text{,} \\ \text{ subject to: }\left\|x-\tilde{x}\right\| \leq \epsilon \text{,}
\end{gathered}
\end{equation}

where $\boldsymbol{\tilde{y}}=\delta(h_{\theta}(\tilde{x}))$, $\mathcal{L}(\cdot)$ denote the specific loss function used by each adversarial attack and $\mathcal{L}_{ce}(\cdot)$ is the cross-entropy loss of the target model, i.e. $\mathcal{L}_{ce}(\boldsymbol{\tilde{y}}, \boldsymbol{y})=-\frac{1}{n} \sum_{i=1}^{n} \boldsymbol{y_i} \cdot \log(\boldsymbol{\tilde{y}})$.
Similarly, we can generate adversarial instances via the target attack. The details of the target attack are presented in Appendix~\ref{appendix_3_3}. 

We combine this dual attack strategy with five attacks to evaluate  our defense model trained in \textit{Scenario (i)}. As shown in Tab.~\ref{tab2}, the results in the second row are the adversarial accuracy of our MAN-based defense. The results in the third row (Model$_T$) are the accuracy of the target model for adversarial instances. The results demonstrate that our defense method can help infer natural labels by using the transition matrix and adversarial labels, and can provide effective protection against multiple attacks in this scenario. 

We note that the adversarial accuracy of MAN in Tab.~\ref{tab2} is higher than that in Tab.~\ref{tab1}. This may be because, for some input instances, the dual attack mainly uses the gradient information of the attack loss against the target model, to generate adversarial noise for breaking the target model. Our MAN-based defense can model the adversarial noise (i.e., learn the transition relationship for the adversarial label predicted by the target model) and infer the final natural label. The final adversarial accuracy thus can be improved.

We just showed three examples for non-targeted and targeted attacks designed to our defense model in the above three scenarios. Note that more different attacks can be designed, which however is beyond the scope of our work in this paper. 

\begin{table}[t]
\caption{Adversarial accuracy (percentage) of our MAN-based defense against dual white-box adaptive attacks on \textit{CIFAR-10}. The target model is ResNet-18.}
\label{tab2}
\renewcommand\tabcolsep{4.5pt}
\renewcommand\arraystretch{1.1}
\begin{center}
\begin{small}
\begin{tabular}{l|cccccc}
\hline
Defense & None & PGD-40 & AA & FWA-40 & CW & DDN \\ \hline
MAN & 82.72 & 70.63 & 45.13 & 64.09  & 45.21 & 38.16 \\ 
Model$_T$ &8.67 &2.36 &1.43 &1.65 &2.10 &0.73 \\ \hline
\end{tabular}
\end{small}
\end{center}
\vskip -0.2in
\end{table}

\begin{table}[t]
\caption{Adversarial accuracy (percentage) of our MAN-based defense against general attacks on \textit{CIFAR-10}. The target model is ResNet-18.}
\label{tab3}
\renewcommand\tabcolsep{4.5pt}
\renewcommand\arraystretch{1.1}
\begin{center}
\begin{small}
\begin{tabular}{l|cccccc}
\hline
Defense & None & PGD-40 & AA & FWA-40 & CW & DDN \\ \hline
MAN &89.01 & 81.07 & 79.90 & 80.02 &77.89 & 77.82 \\ 
Model$_T$ &88.98 &0.00 &0.00 &0.00 &0.00 &0.00 \\ \hline
\end{tabular}
\end{small}
\end{center}
\vskip -0.2in
\end{table}

\textbf{Defending against general attacks.}
We also evaluate the effectiveness of the proposed defense method against general adversarial attacks. The general adversarial attacks usually only focus on disrupting the performance of the target model. We utilize five adversarial attacks to evaluate the proposed MAN-based defense method. To train the defense model, we use the adversarial instances crafted by non-target $L_{\infty}$-norm PGD-10 attack against the target model as the adversarial training data. The adversarial accuracy of our MAN-based defense and the adversarial accuracy of the target model (Model$_T$) are shown in Tab.~\ref{tab3}. It can be seen that our defense method achieve a great defense effect against general adversarial attacks. Note that we only train the defense model in the proposed method (i.e., the transition network) and fix the model parameters of the target model in this experiment. In addition, to illustrate that our method does not utilize gradient masking, we conduct additional experiments listed in \citet{athalye2018obfuscated}. Detailed results can be found in the Appendix~\ref{appendix_4}

\textbf{Defense transferability.}
The work in \citet{ilyas2019adversarial} shows that any two target models are likely to learn similar mixture pattern. Therefore, modeling the adversarial noise which manipulate such features would apply to both target models for improving adversarial accuracy. We apply the MAN-based defense which is trained for the ResNet-18 (ResNet), to other naturally pre-trained target models, such as VGG-19 (VGG) \cite{simonyan2014very} and Wide-ResNet 28$\times$10 (WRN) \cite{Zagoruyko2016WRN} for evaluating the transferability of our defense method. Against general adversarial attacks, we deploy the defense model trained in Section \textit{Defending against general attacks} on the VGG and WRN target models. In addition, against adaptive adversarial attacks, we deploy our defense model trained in Section \textit{Scenario (i)} on VGG. 

\begin{table}[t]
\caption{Adversarial accuracy (percentage) of our defense method for different target models on \textit{CIFAR-10}. MAN$^{\star}$ denote the fine-tuned defense model on VGG.}
\label{tab4}
\renewcommand\tabcolsep{4.7pt}
\renewcommand\arraystretch{0.9}
\begin{center}
\begin{small}
\begin{tabular}{l|ccccc}
\hline
Defense & PGD-40 & AA & FWA-40 & CW$_2$ & DDN \\ \hline
\multicolumn{6}{c}{General adverarial attacks} \\ 
ResNet-MAN & 81.07 &79.90 & 80.02 & 77.89 & 77.82 \\
VGG-MAN & 71.93 & 68.62 & 68.19 & 70.91 & 70.76 \\
WRN-MAN & 72.08 & 69.13 & 68.83 & 71.12 & 70.93 \\ \hline
\multicolumn{6}{c}{Adaptive adverarial attacks} \\
ResNet-MAN & 44.83 & 39.43 & 29.53 & 43.17 & 10.63 \\
VGG-MAN & 11.71 & 7.36 & 6.98 & 6.31 & 2.29 \\ 
VGG-MAN$^{\star}$ & 33.26 & 30.30 & 13.42 & 10.07 & 3.83 \\ \hline
\end{tabular}
\end{small}
\end{center}
\vskip -0.1in
\end{table}

The performances on \textit{CIFAR-10} are reported in Tab.~\ref{tab4}. It can be seen that our defense method has a certain degree of transferability for providing cross-model protections against adversarial attacks. Moreover, for black-box target models whose model parameters and training manners cannot be accessed (e.g., the VGG target model), we can fine-tune the transition network in our defense method to further improve the defense effect against adaptive adversarial attacks. As shown by MAN$^{\star}$ in Tab.~\ref{tab4}, the fine-tuned defense model achieves a higher adversarial accuracy. 

\subsection{Ablation study}
\label{section4.3}
\begin{figure}[t]
\vskip 0.1in
    \centering
    \label{fig3}
    \includegraphics[width=2.0 in]{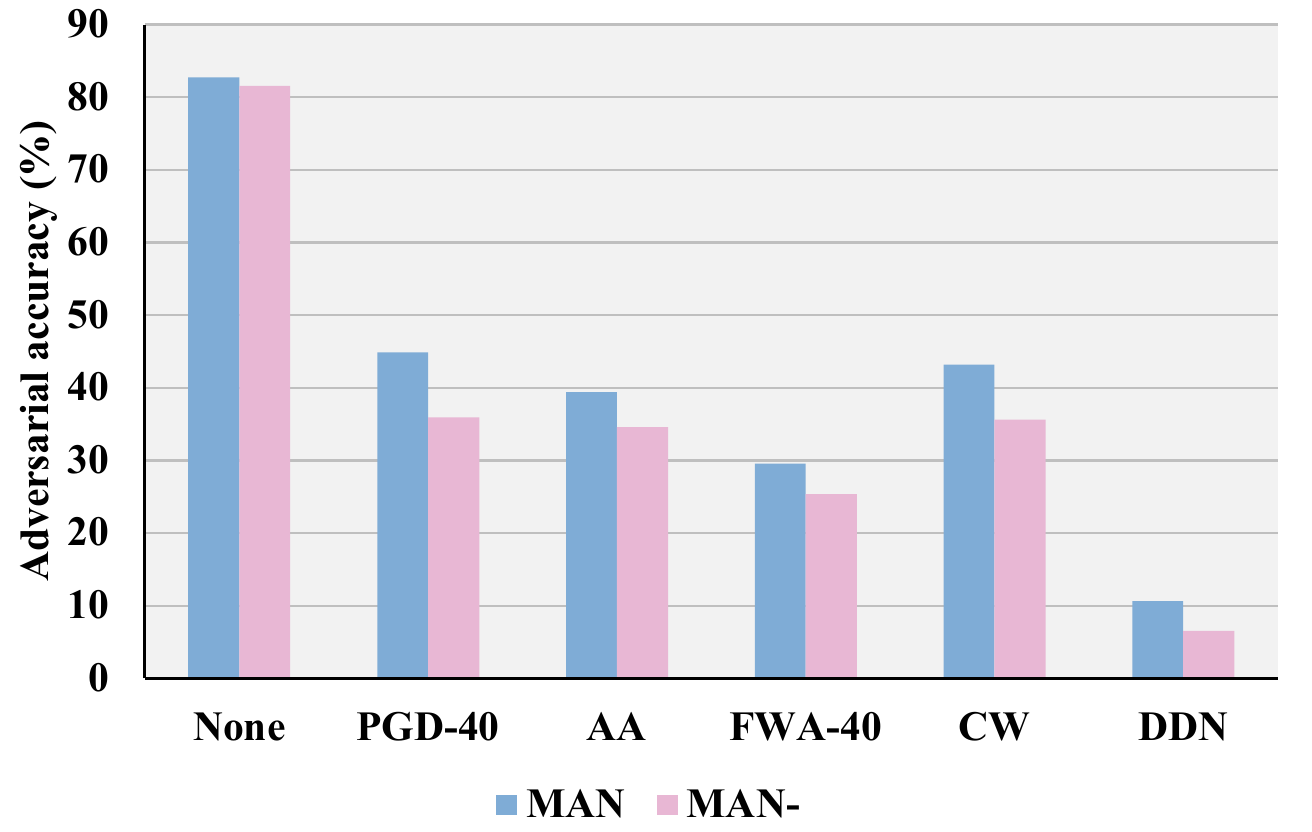}
    \vskip -0.1in
    \caption{Ablation study by independently training the transition network.}
\vskip -0.1in
\end{figure}

To demonstrate that compared with only training the transition network, the joint adversarial training on the target model and the transition network could achieve better defense effectiveness, we conduct an ablation study. Specifically, we use a naturally pre-trained ResNet-18 target model, and train the transition network independently (called MAN-). The model parameters of the target model are fixed in the training procedure. We use the adversarial instances crafted by the adaptive $L_{\infty}$-norm PGD-10 as adversarial training data. Compared with jointly trained MAN, the results shown in Fig.~\ref{fig3} demonstrated the joint adversarial training manner can provide positive gains. 

In addition, to demonstrate that the improvement of our method has little relationship with the fact that the transition network introduces more model parameters, we conduct an experiment by using two parallel ResNet-18 as the target model of the baseline. We use the AT method to train the new target model. The result demonstrates that the gain of our method is not due to the increase of model parameters. The details and results can be found in Appendix~\ref{appendix_5}.

\subsection{Detecting adversarial samples}
\label{section4.4}
\begin{figure}[htbp]
\centering
\subfigure[PGD]{\includegraphics[height=2.6cm,width=2.6cm]{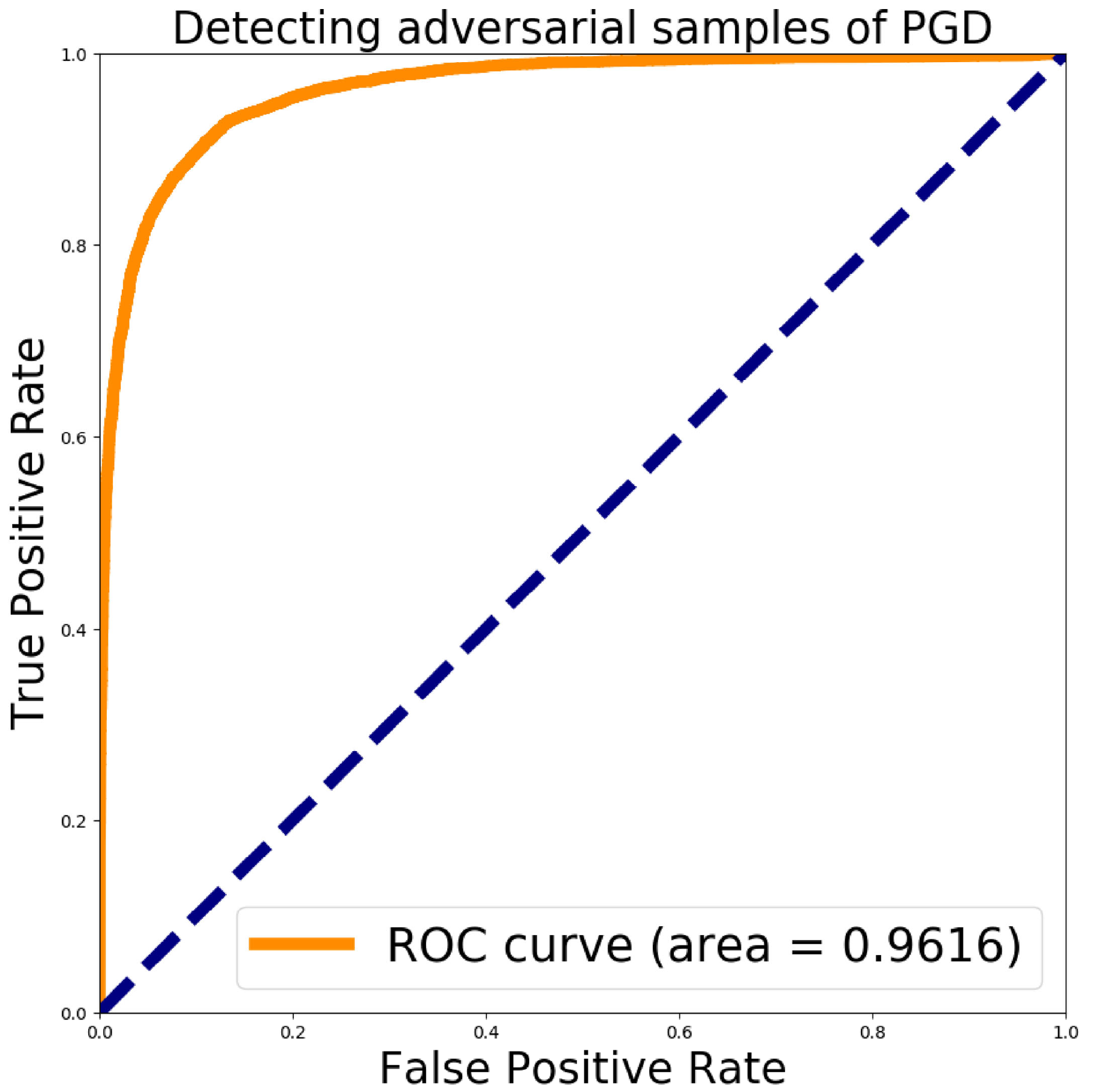}}
\hspace{0.30in}
\subfigure[DDN]{\includegraphics[height=2.6cm,width=2.6cm]{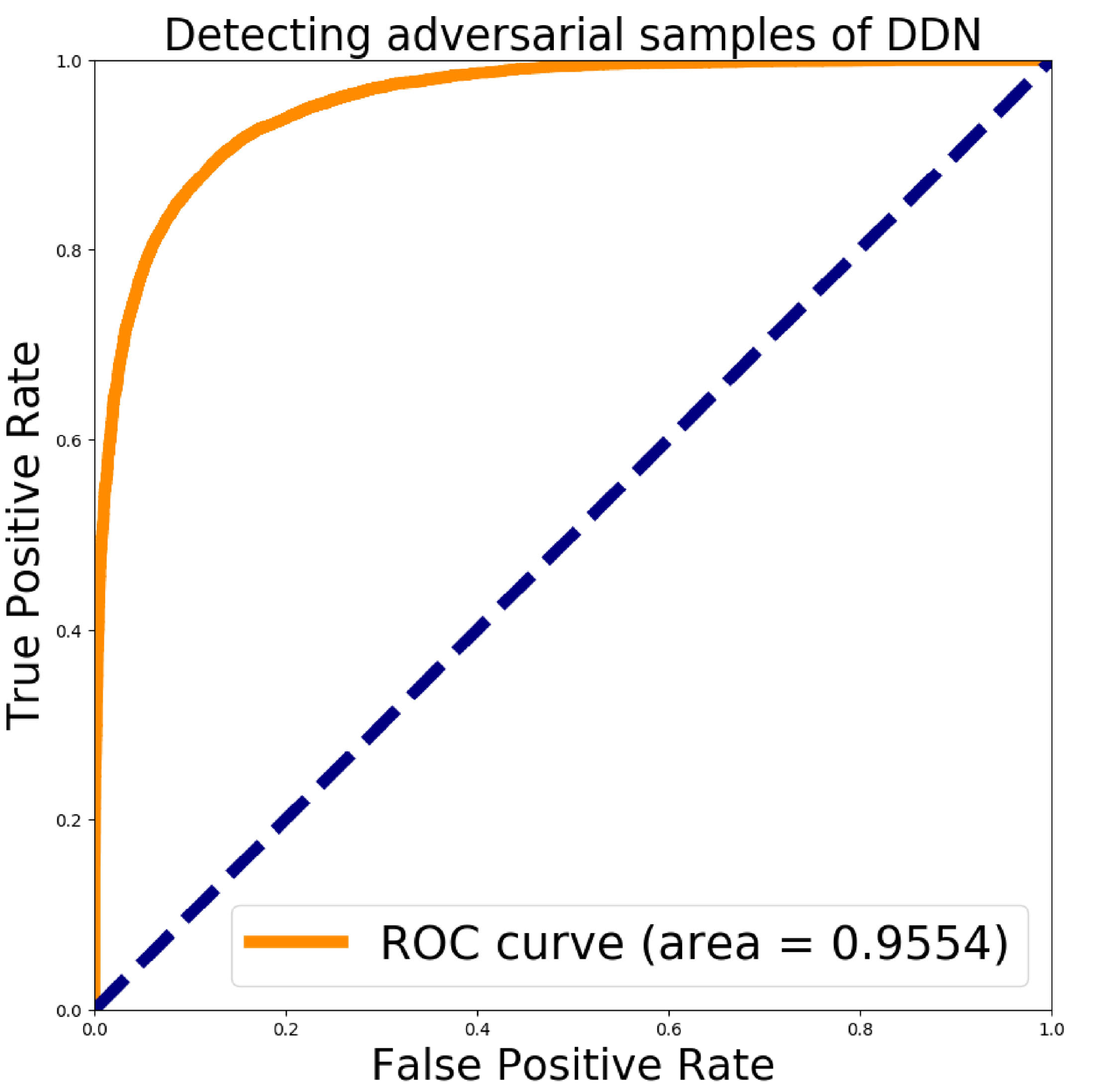}}
\vskip -0.1in
\caption{AUROC of detecting adversarial samples.}
\label{fig4}
\end{figure}

Besides improving adversarial robustness, we find that the proposed MAN can be utilized to detect adversarial samples. By observing the distribution of element values on the diagonal of the generated transition matrix, we can discriminate whether the input is natural or adversarial. Specifically, we use the label predicted by the target model as the index to obtain the element value $\boldsymbol{p}$ from the diagonal of the transition matrix. If the value of the obtained element is larger than the values of other elements on the diagonal, the input sample is a natural sample, otherwise, it is an adversarial sample. Therefore, we take $1-\boldsymbol{p}$ as the probability that the input is adversarial. We use 10,000 test samples on \textit{CIFAR-10} to compute the AUROC against $L_{\infty}$-norm PGD-40 and $L_{2}$-norm DDN (see Fig.~\ref{fig4}). These results further validate the roles of the transition network and the transition matrix in defending against adversarial attacks.

\section{Conclusion}
\label{section5}
Traditional adversarial defense methods typically focus on directly exploiting adversarial instances to remove adversarial noise or train an adversarially robust model. In this paper, motivated by that the relationship between adversarial data and natural data can help infer clean data from adversarial data, we study to model adversarial noise by learning the label transition relationship for improving adversarial accuracy. We propose a defense method based on Modeling Adversarial Noise (called MAN). Specifically, we embed a label transition matrix into the target model, which denote the transition relationship from adversarial labels to natural labels. The transition matrix explicitly models adversarial noise and help us infer natural labels. We design a transition network to generate the instance-independent transition matrix. Considering that adversarial data can be adaptively generated, we conduct joint adversarial training on the target model and the transition network to achieve an optimal adversarial accuracy. The empirical results demonstrate that our defense method can provide effective protection against white-box general attacks and adaptive attacks. Our work provides a new adversarial defense strategy for the community of adversarial learning. In future, we will further optimize the MAN-based defense method to improve its transferability and its performance when applied to other adversarial training methods. For datasets with more classes, how to effectively learn the transition matrix is also our future focus.

\section{Acknowledgements}
This work was supported in part by the National Key Research and Development Program of China under Grant 2018AAA0103202, in part by the National Natural Science Foundation of China under Grant 61922066, 61876142, 62036007 and 62006202, in part by the Technology Innovation Leading Program of Shaanxi under Grant 2022QFY01-15, in part by Open Research Projects of Zhejiang Lab under Grant 2021KG0AB01, in part by the RGC Early Career Scheme No. 22200720, in part by Guangdong Basic and Applied Basic Research Foundation No. 2022A1515011652, in part by Australian Research Council Projects DE-190101473, IC-190100031, and DP-220102121, in part by the Fundamental Research Funds for the Central Universities, and in part by the Innovation Fund of Xidian University. The authors thank the reviewers and the meta-reviewer for their helpful and constructive comments on this work. Thanks to Chaojian Yu for his important advice on Section \textit{attack the transition matrix}.

\bibliography{example_paper}

\begin{thebibliography}{38}
\providecommand{\natexlab}[1]{#1}
\providecommand{\url}[1]{\texttt{#1}}
\expandafter\ifx\csname urlstyle\endcsname\relax
  \providecommand{\doi}[1]{doi: #1}\else
  \providecommand{\doi}{doi: \begingroup \urlstyle{rm}\Url}\fi

\bibitem[Athalye et~al.(2018)Athalye, Carlini, and
  Wagner]{athalye2018obfuscated}
Athalye, A., Carlini, N., and Wagner, D.~A.
\newblock Obfuscated gradients give a false sense of security: Circumventing
  defenses to adversarial examples.
\newblock In \emph{Proceedings of the 35th International Conference on Machine
  Learning}, 2018.

\bibitem[Carlini \& Wagner(2017{\natexlab{a}})Carlini and
  Wagner]{carlini2017magnet}
Carlini, N. and Wagner, D.
\newblock Magnet and" efficient defenses against adversarial attacks" are not
  robust to adversarial examples.
\newblock \emph{arXiv preprint arXiv:1711.08478}, 2017{\natexlab{a}}.

\bibitem[Carlini \& Wagner(2017{\natexlab{b}})Carlini and
  Wagner]{carlini2017towards}
Carlini, N. and Wagner, D.
\newblock Towards evaluating the robustness of neural networks.
\newblock In \emph{2017 Ieee Symposium on Security and Privacy (sp)}, pp.\
  39--57. IEEE, 2017{\natexlab{b}}.

\bibitem[Carmon et~al.(2019)Carmon, Raghunathan, Schmidt, Duchi, and
  Liang]{carmon2019unlabeled}
Carmon, Y., Raghunathan, A., Schmidt, L., Duchi, J.~C., and Liang, P.
\newblock Unlabeled data improves adversarial robustness.
\newblock In Wallach, H.~M., Larochelle, H., Beygelzimer, A.,
  d'Alch{\'{e}}{-}Buc, F., Fox, E.~B., and Garnett, R. (eds.), \emph{Advances
  in Neural Information Processing Systems 32: Annual Conference on Neural
  Information Processing Systems 2019, NeurIPS 2019, December 8-14, 2019,
  Vancouver, BC, Canada}, pp.\  11190--11201, 2019.

\bibitem[Croce \& Hein(2020)Croce and Hein]{croce2020reliable}
Croce, F. and Hein, M.
\newblock Reliable evaluation of adversarial robustness with an ensemble of
  diverse parameter-free attacks.
\newblock In \emph{Proceedings of the 37th International Conference on Machine
  Learning}, 2020.

\bibitem[Ding et~al.(2019)Ding, Lui, Jin, Wang, and Huang]{ding2019sensitivity}
Ding, G.~W., Lui, K. Y.~C., Jin, X., Wang, L., and Huang, R.
\newblock On the sensitivity of adversarial robustness to input data
  distributions.
\newblock In \emph{ICLR (Poster)}, 2019.

\bibitem[Goodfellow et~al.(2015)Goodfellow, Shlens, and
  Szegedy]{goodfellow2014explaining}
Goodfellow, I.~J., Shlens, J., and Szegedy, C.
\newblock Explaining and harnessing adversarial examples.
\newblock In \emph{International Conference on Learning Representations}, 2015.

\bibitem[Guo et~al.(2018)Guo, Rana, Ciss{\'{e}}, and van~der
  Maaten]{guo2017countering}
Guo, C., Rana, M., Ciss{\'{e}}, M., and van~der Maaten, L.
\newblock Countering adversarial images using input transformations.
\newblock In \emph{6th International Conference on Learning Representations,
  {ICLR} 2018, Vancouver, BC, Canada, April 30 - May 3, 2018, Conference Track
  Proceedings}. OpenReview.net, 2018.

\bibitem[He et~al.(2016)He, Zhang, Ren, and Sun]{he2016deep}
He, K., Zhang, X., Ren, S., and Sun, J.
\newblock Deep residual learning for image recognition.
\newblock In \emph{Conference on Computer Vision and Pattern Recognition}, pp.\
   770--778, 2016.

\bibitem[Ilyas et~al.(2019)Ilyas, Santurkar, Tsipras, Engstrom, Tran, and
  Madry]{ilyas2019adversarial}
Ilyas, A., Santurkar, S., Tsipras, D., Engstrom, L., Tran, B., and Madry, A.
\newblock Adversarial examples are not bugs, they are features.
\newblock \emph{arXiv preprint arXiv:1905.02175}, 2019.

\bibitem[Jin et~al.(2019)Jin, Shen, Zhang, Dai, and Zhang]{shen2017ape}
Jin, G., Shen, S., Zhang, D., Dai, F., and Zhang, Y.
\newblock {APE-GAN:} adversarial perturbation elimination with {GAN}.
\newblock In \emph{International Conference on Acoustics, Speech and Signal
  Processing}, pp.\  3842--3846, 2019.

\bibitem[Kaiming et~al.(2017)Kaiming, Georgia, Piotr, and Ross]{2017Mask}
Kaiming, H., Georgia, G., Piotr, D., and Ross, G.
\newblock Mask r-cnn.
\newblock \emph{IEEE Transactions on Pattern Analysis \& Machine Intelligence},
  PP:\penalty0 1--1, 2017.

\bibitem[Krizhevsky et~al.(2009)Krizhevsky, Hinton,
  et~al.]{krizhevsky2009learning}
Krizhevsky, A., Hinton, G., et~al.
\newblock Learning multiple layers of features from tiny images.
\newblock 2009.

\bibitem[LeCun et~al.(1998)LeCun, Bottou, Bengio, and
  Haffner]{lecun1998gradient}
LeCun, Y., Bottou, L., Bengio, Y., and Haffner, P.
\newblock Gradient-based learning applied to document recognition.
\newblock \emph{Proceedings of the IEEE}, 86\penalty0 (11):\penalty0
  2278--2324, 1998.

\bibitem[Liao et~al.(2018)Liao, Liang, Dong, Pang, Hu, and
  Zhu]{liao2018defense}
Liao, F., Liang, M., Dong, Y., Pang, T., Hu, X., and Zhu, J.
\newblock Defense against adversarial attacks using high-level representation
  guided denoiser.
\newblock In \emph{Conference on Computer Vision and Pattern Recognition}, pp.\
   1778--1787, 2018.

\bibitem[Liu \& Tao(2015)Liu and Tao]{liu2015classification}
Liu, T. and Tao, D.
\newblock Classification with noisy labels by importance reweighting.
\newblock \emph{IEEE Transactions on pattern analysis and machine
  intelligence}, 38\penalty0 (3):\penalty0 447--461, 2015.

\bibitem[Ma et~al.(2018)Ma, Li, Wang, Erfani, Wijewickrema, Schoenebeck, Song,
  Houle, and Bailey]{ma2018characterizing}
Ma, X., Li, B., Wang, Y., Erfani, S.~M., Wijewickrema, S. N.~R., Schoenebeck,
  G., Song, D., Houle, M.~E., and Bailey, J.
\newblock Characterizing adversarial subspaces using local intrinsic
  dimensionality.
\newblock In \emph{International Conference on Learning Representations}, 2018.

\bibitem[Ma et~al.(2021)Ma, Niu, Gu, Wang, Zhao, Bailey, and
  Lu]{ma2021understanding}
Ma, X., Niu, Y., Gu, L., Wang, Y., Zhao, Y., Bailey, J., and Lu, F.
\newblock Understanding adversarial attacks on deep learning based medical
  image analysis systems.
\newblock \emph{Pattern Recognition}, 110:\penalty0 107332, 2021.

\bibitem[Madry et~al.(2018)Madry, Makelov, Schmidt, Tsipras, and
  Vladu]{madry2017towards}
Madry, A., Makelov, A., Schmidt, L., Tsipras, D., and Vladu, A.
\newblock Towards deep learning models resistant to adversarial attacks.
\newblock In \emph{6th International Conference on Learning Representations},
  2018.

\bibitem[Naseer et~al.(2020)Naseer, Khan, Hayat, Khan, and
  Porikli]{naseer2020self}
Naseer, M., Khan, S., Hayat, M., Khan, F.~S., and Porikli, F.
\newblock A self-supervised approach for adversarial robustness.
\newblock In \emph{Proceedings of the IEEE/CVF Conference on Computer Vision
  and Pattern Recognition}, pp.\  262--271, 2020.

\bibitem[Rony et~al.(2019)Rony, Hafemann, Oliveira, Ayed, Sabourin, and
  Granger]{rony2019decoupling}
Rony, J., Hafemann, L.~G., Oliveira, L.~S., Ayed, I.~B., Sabourin, R., and
  Granger, E.
\newblock Decoupling direction and norm for efficient gradient-based {L2}
  adversarial attacks and defenses.
\newblock In \emph{Conference on Computer Vision and Pattern Recognition}, pp.\
   4322--4330, 2019.

\bibitem[Simonyan \& Zisserman(2015)Simonyan and Zisserman]{simonyan2014very}
Simonyan, K. and Zisserman, A.
\newblock Very deep convolutional networks for large-scale image recognition.
\newblock In Bengio, Y. and LeCun, Y. (eds.), \emph{3rd International
  Conference on Learning Representations}, 2015.

\bibitem[Szegedy et~al.(2014)Szegedy, Zaremba, Sutskever, Bruna, Erhan,
  Goodfellow, and Fergus]{szegedy2013intriguing}
Szegedy, C., Zaremba, W., Sutskever, I., Bruna, J., Erhan, D., Goodfellow,
  I.~J., and Fergus, R.
\newblock Intriguing properties of neural networks.
\newblock In \emph{International Conference on Learning Representations}, 2014.

\bibitem[Wang et~al.(2019)Wang, Zou, Yi, Bailey, Ma, and Gu]{wang2019improving}
Wang, Y., Zou, D., Yi, J., Bailey, J., Ma, X., and Gu, Q.
\newblock Improving adversarial robustness requires revisiting misclassified
  examples.
\newblock In \emph{International Conference on Learning Representations}, 2019.

\bibitem[Wu et~al.(2020{\natexlab{a}})Wu, Xia, and Wang]{wu2020adversarial}
Wu, D., Xia, S.-T., and Wang, Y.
\newblock Adversarial weight perturbation helps robust generalization.
\newblock \emph{Advances in Neural Information Processing Systems}, 33,
  2020{\natexlab{a}}.

\bibitem[Wu et~al.(2017)Wu, Zhang, and Xu]{wu2017tiny}
Wu, J., Zhang, Q., and Xu, G.
\newblock Tiny imagenet challenge.
\newblock \emph{Technical Report}, 2017.

\bibitem[Wu et~al.(2020{\natexlab{b}})Wu, Wang, and Yu]{wu2020stronger}
Wu, K., Wang, A.~H., and Yu, Y.
\newblock Stronger and faster wasserstein adversarial attacks.
\newblock In \emph{Proceedings of the 37th International Conference on Machine
  Learning}, volume 119, pp.\  10377--10387, 2020{\natexlab{b}}.

\bibitem[Wu et~al.(2021)Wu, Xia, Liu, Han, Gong, Wang, Liu, and
  Niu]{wu2021class2simi}
Wu, S., Xia, X., Liu, T., Han, B., Gong, M., Wang, N., Liu, H., and Niu, G.
\newblock Class2simi: A noise reduction perspective on learning with noisy
  labels.
\newblock In \emph{International Conference on Machine Learning}, pp.\
  11285--11295. PMLR, 2021.

\bibitem[Xia et~al.(2019)Xia, Liu, Wang, Han, Gong, Niu, and
  Sugiyama]{xia2019anchor}
Xia, X., Liu, T., Wang, N., Han, B., Gong, C., Niu, G., and Sugiyama, M.
\newblock Are anchor points really indispensable in label-noise learning?
\newblock \emph{arXiv preprint arXiv:1906.00189}, 2019.

\bibitem[Xia et~al.(2020)Xia, Liu, Han, Wang, Gong, Liu, Niu, Tao, and
  Sugiyama]{xia2020part}
Xia, X., Liu, T., Han, B., Wang, N., Gong, M., Liu, H., Niu, G., Tao, D., and
  Sugiyama, M.
\newblock Part-dependent label noise: Towards instance-dependent label noise.
\newblock \emph{Advances in Neural Information Processing Systems}, 33, 2020.

\bibitem[Xia et~al.(2021)Xia, Liu, Han, Gong, Wang, Ge, and
  Chang]{xia2021robust}
Xia, X., Liu, T., Han, B., Gong, C., Wang, N., Ge, Z., and Chang, Y.
\newblock Robust early-learning: Hindering the memorization of noisy labels.
\newblock In \emph{International Conference on Learning Representations}, 2021.

\bibitem[Xiao et~al.(2018)Xiao, Zhu, Li, He, Liu, and Song]{xiao2018spatially}
Xiao, C., Zhu, J., Li, B., He, W., Liu, M., and Song, D.
\newblock Spatially transformed adversarial examples.
\newblock In \emph{6th International Conference on Learning Representations},
  2018.

\bibitem[Xu et~al.(2017)Xu, Evans, and Qi]{xu2017feature}
Xu, W., Evans, D., and Qi, Y.
\newblock Feature squeezing: Detecting adversarial examples in deep neural
  networks.
\newblock \emph{arXiv preprint arXiv:1704.01155}, 2017.

\bibitem[Yang et~al.(2021)Yang, Yang, Han, Liu, Xu, Niu, and
  Liu]{yang2021estimating}
Yang, S., Yang, E., Han, B., Liu, Y., Xu, M., Niu, G., and Liu, T.
\newblock Estimating instance-dependent label-noise transition matrix using
  dnns.
\newblock \emph{arXiv preprint arXiv:2105.13001}, 2021.

\bibitem[Zagoruyko \& Komodakis(2016)Zagoruyko and Komodakis]{Zagoruyko2016WRN}
Zagoruyko, S. and Komodakis, N.
\newblock Wide residual networks.
\newblock In Wilson, R.~C., Hancock, E.~R., and Smith, W. A.~P. (eds.),
  \emph{Proceedings of the British Machine Vision Conference 2016}, 2016.

\bibitem[Zhang et~al.(2019)Zhang, Yu, Jiao, Xing, El~Ghaoui, and
  Jordan]{zhang2019theoretically}
Zhang, H., Yu, Y., Jiao, J., Xing, E., El~Ghaoui, L., and Jordan, M.
\newblock Theoretically principled trade-off between robustness and accuracy.
\newblock In \emph{International Conference on Machine Learning}, pp.\
  7472--7482. PMLR, 2019.

\bibitem[Zhou et~al.(2021{\natexlab{a}})Zhou, Liu, Han, Wang, Peng, and
  Gao]{pmlr-v139-zhou21e}
Zhou, D., Liu, T., Han, B., Wang, N., Peng, C., and Gao, X.
\newblock Towards defending against adversarial examples via attack-invariant
  features.
\newblock In \emph{Proceedings of the 38th International Conference on Machine
  Learning}, pp.\  12835--12845, 2021{\natexlab{a}}.

\bibitem[Zhou et~al.(2021{\natexlab{b}})Zhou, Wang, Peng, Gao, Wang, Yu, and
  Liu]{zhou2021removing}
Zhou, D., Wang, N., Peng, C., Gao, X., Wang, X., Yu, J., and Liu, T.
\newblock Removing adversarial noise in class activation feature space.
\newblock In \emph{Proceedings of the IEEE/CVF International Conference on
  Computer Vision}, pp.\  7878--7887, 2021{\natexlab{b}}.

\end{thebibliography}
\bibliographystyle{icml2022}

\newpage
\appendix
\onecolumn
\large{Appendices}
\normalsize

\section{The difference of the transition matrix between our method and label-noise learning}
\label{appendix_1}
In label-noise learning, the transition matrix is used to infer clean labels from given noisy labels. The transition matrix denotes the probabilities that \textbf{clean} labels flip into \textbf{noisy} labels \cite{xia2020part,liu2015classification,yang2021estimating}. The label-noise learning methods utilize the transition matrix to correct the training loss on noisy data (i.e., the clean instance with the \textbf{noisy} label). Given the noisy class posterior probability, the clean class posterior probability can be obtained, i.e., $p(\boldsymbol{y}|x)=(T(x)^{\top})^{-1} p(\boldsymbol{y}^{*}|x)$, where $\boldsymbol{y}^{*}$ denotes the noisy label in the form of a vector. Differently, in our method, the transition matrix is used to infer natural labels from observed adversarial labels. The transition matrix denotes the probabilities that \textbf{adversarial} labels flip into \textbf{natural} labels. We utilize the transition matrix to help compute the training loss on the adversarial instance with the \textbf{natural} label. For the observed adversarial class posterior probability, the natural class posterior probability can be obtained, i.e., $p(\boldsymbol{y}|\tilde{x})=T(\tilde{x})^{\top}p(\tilde{\boldsymbol{y}}|\tilde{x})$. In addition, to the best of our knowledge, our method for the first time utilizes the transition matrix to explicitly model adversarial noise for improving adversarial robustness.

\section{Training settings}
\label{appendix_2}
For all baselines and our defense method, we use the $L_{\infty}$-norm non-target PGD-10 (i.e., PGD with iteration number of 10) with random start and step size $\epsilon/4$ to craft adversarial training data. The perturbation budget $\epsilon$ is set to $8/255$ for both \textit{CIFAR-10} and \textit{Tiny-ImageNet}. All the defense models are trained using SGD with momentum 0.9 and an initial learning rate of 0.1. 
The weight decay is $2 \times 10^{-4}$ for \textit{CIFAR-10}, and is $5 \times 10^{-4}$ for \textit{Tiny-ImageNet}. The batch-size is set as 1024 to reduce time cost. For a fair comparison, we adjust the hyperparameter settings of the defense methods so that the natural accuracy is not severely compromised and then compare the robustness. 
The epoch number is set to 100. The learning rate is divided by 10 at the 75-th and 90-th epoch. We report the evaluation results of the \textit{last} checkpoint rather than those of the \textit{best} checkpoint.

\section{Defending against adaptive attacks}
\label{appendix_3}

\subsection{Scenario (i): disturb the final output}
\label{appendix_3_1}

\begin{table*}[hbtp]
\vskip -0.1in
\caption{Adversarial accuracy (percentage) of defense methods against white-box adaptive attacks on \textit{CIFAR-10}. The target model is ResNet-18. We show the most successful defense with \textbf{bold}.}
\label{appendix_tab1}
\renewcommand\tabcolsep{4.0 pt}
\renewcommand\arraystretch{1.2}
\begin{center}
\begin{tabular}{l|cccccc}
\hline
Defense & None & PGD-40 & AA & FWA-40 & CW$_2$ & DDN \\ \hline
AT & \textbf{83.39$\pm$0.95} & 42.38$\pm$0.56& 39.01$\pm$0.51& 15.44$\pm$0.32 &0.00$\pm$0.00 & 0.09$\pm$0.03 \\
MAN & 82.72$\pm$0.53 & \textbf{44.83$\pm$0.47} & \textbf{39.43$\pm$0.73} & \textbf{29.53$\pm$0.47} &\textbf{43.17$\pm$0.68} & \textbf{10.63$\pm$0.49} \\ \cdashline{1-7}[4pt/7pt]
TRADES & \textbf{80.70$\pm$0.63} & 46.29$\pm$0.59 & 42.71$\pm$0.49 & 20.54$\pm$0.47 &0.00$\pm$0.00 & 0.06$\pm$0.01 \\
MAN\_TRADES &80.34$\pm$0.61  &\textbf{48.65$\pm$0.41}  &\textbf{44.40$\pm$0.56}  &\textbf{29.13$\pm$0.60} &\textbf{1.46$\pm$0.21} &\textbf{0.31$\pm$0.05} \\ \cdashline{1-7}[4pt/7pt]
MART & \textbf{78.21$\pm$0.65} & 50.23$\pm$0.70 & 43.96$\pm$0.67 & 25.56$\pm$0.61 &0.02$\pm$0.00 & 0.07$\pm$0.01  \\
MAN\_MART &77.83$\pm$0.67  &\textbf{50.95$\pm$0.61}  &\textbf{44.42$\pm$0.69}  &\textbf{31.23$\pm$0.58} &\textbf{1.53$\pm$0.27} &\textbf{0.47$\pm$0.07} \\ \hline
\end{tabular}
\vskip -0.1in
\end{center}
\end{table*}

\begin{table*}[hbtp]
\vskip -0.1in
\caption{Adversarial accuracy (percentage) of defense methods against white-box adaptive attacks on \textit{Tiny-ImageNet}. The target model is ResNet-18. We show the most successful defense with \textbf{bold}.}
\label{appendix_tab2}
\renewcommand\tabcolsep{4.0 pt}
\renewcommand\arraystretch{1.2}
\begin{center}
\begin{tabular}{l|cccccc}
\hline
Defense & None & PGD-40 & AA & FWA-40 & CW$_2$ &DDN \\ \hline
AT & \textbf{48.40$\pm$0.68} & 17.35$\pm$0.59& 11.27$\pm$0.53 & 10.29$\pm$0.47  & 0.00$\pm$0.00 & 0.29$\pm$0.03 \\
MAN & 48.29$\pm$0.57 & \textbf{18.15$\pm$0.51} & \textbf{12.45$\pm$0.67} & \textbf{13.17$\pm$0.69} & \textbf{16.27$\pm$0.71} & \textbf{4.01$\pm$0.19} \\ \cdashline{1-7}[4pt/7pt]
TRADES & \textbf{48.25$\pm$0.71} & 19.17$\pm$0.58 & 12.63$\pm$0.51 & 10.67$\pm$0.68 & 0.00$\pm$0.00 & 0.05$\pm$0.01 \\
MAN\_TRADES &48.19$\pm$0.62  &\textbf{20.12$\pm$0.49}  &\textbf{12.86$\pm$0.59}  & \textbf{14.91$\pm$0.47}  &\textbf{0.67$\pm$0.12} &\textbf{1.10$\pm$0.15} \\ \cdashline{1-7}[4pt/7pt]
MART & \textbf{47.83$\pm$0.65} & 20.90$\pm$0.59 & 15.57$\pm$0.52 & 12.95$\pm$0.49 & 0.00$\pm$0.00 & 0.06$\pm$0.01  \\
MAN\_MART &47.79$\pm$0.59 &\textbf{21.22$\pm$0.60}  &\textbf{15.84$\pm$0.47}  &\textbf{15.10$\pm$0.39} &\textbf{0.89$\pm$0.16} &\textbf{1.23$\pm$0.21} \\ \hline
\end{tabular}
\end{center}
\end{table*}

In scenario (i), the target attack solve the following optimization problem: 
\begin{equation}
\label{eq9}
\max _{\tilde{x}} \, -\mathcal{L}(\boldsymbol{\tilde{y}} \cdot \widehat{T}(\tilde{x};\omega), \boldsymbol{y^{*}}) \text{,} \, \text{ subject to: }\left\|x-\tilde{x}\right\| \leq \epsilon \text{,}
\end{equation}
where $\boldsymbol{y^{*}}$ is the target label in the form of a vector set by the attacker. The adversarial accuracy of defense methods against white-box adaptive attacks on \textit{CIFAR-10} and \textit{Tiny-ImageNet} are shown in Tab.~\ref{appendix_tab1} and Tab.~\ref{appendix_tab2} respectively. 

In addition, we evaluate the adversarial accuracy of defense methods against white-box adaptive attacks on \textit{CIFAR-10} by using the VggNet-19 as the target model and the transition network. As shown in Tab.~
\ref{appendix_tab3}, our method still achieves better performance.

Moreover, we evaluate the robustness performance of our defense method at a small batch-size, such as 128. The initial learning rate is still 0.1. The results are shown in Tab.~\ref{appendix_tab4}.

\begin{table*}[hbtp]
\caption{Adversarial accuracy (percentage) of defense methods against white-box adaptive attacks on \textit{CIFAR-10}. The target model is VggNet-19.}
\label{appendix_tab3}
\renewcommand\tabcolsep{4.0 pt}
\renewcommand\arraystretch{1.2}
\begin{center}
\begin{tabular}{l|cccccc}
\hline
Defense & None & PGD-40 & AA & FWA-40 & CW$_2$ & DDN \\ \hline
AT &\textbf{80.91$\pm$0.61} &29.83$\pm$0.43  &26.00$\pm$0.31 &7.55$\pm$0.19 &0.10$\pm$0.01 &0.15$\pm$0.02 \\
MAN &80.25$\pm$0.50  &\textbf{37.13$\pm$0.61}  &\textbf{33.19$\pm$0.43}  &\textbf{17.88$\pm$0.44}  &\textbf{38.13$\pm$0.37} &\textbf{6.10$\pm$0.09}  \\ \hline
\end{tabular}
\vskip -0.2in
\end{center}
\end{table*}

\begin{table*}[hbtp]
\caption{Adversarial accuracy (percentage) of defense methods against white-box adaptive attacks on \textit{CIFAR-10}. The target model is ResNet-18. The batch-size is 128.}
\label{appendix_tab4}
\renewcommand\tabcolsep{4.0 pt}
\renewcommand\arraystretch{1.2}
\begin{center}
\begin{tabular}{l|cccc}
\hline
Defense & None & PGD-40 & AA \\ \hline
AT &\textbf{84.92$\pm$0.49} &46.47$\pm$0.47  &43.55$\pm$0.43 \\
MAN &84.70$\pm$0.41  &\textbf{48.06$\pm$0.50} &\textbf{44.67$\pm$0.51}\\ \hline
\end{tabular}
\vskip -0.1in
\end{center}
\end{table*}

\subsection{Scenario (ii): attack the transition matrix}
\label{appendix_3_2}

\begin{figure*}[hbtp]
    \centering    
    \includegraphics[width=1.7in]{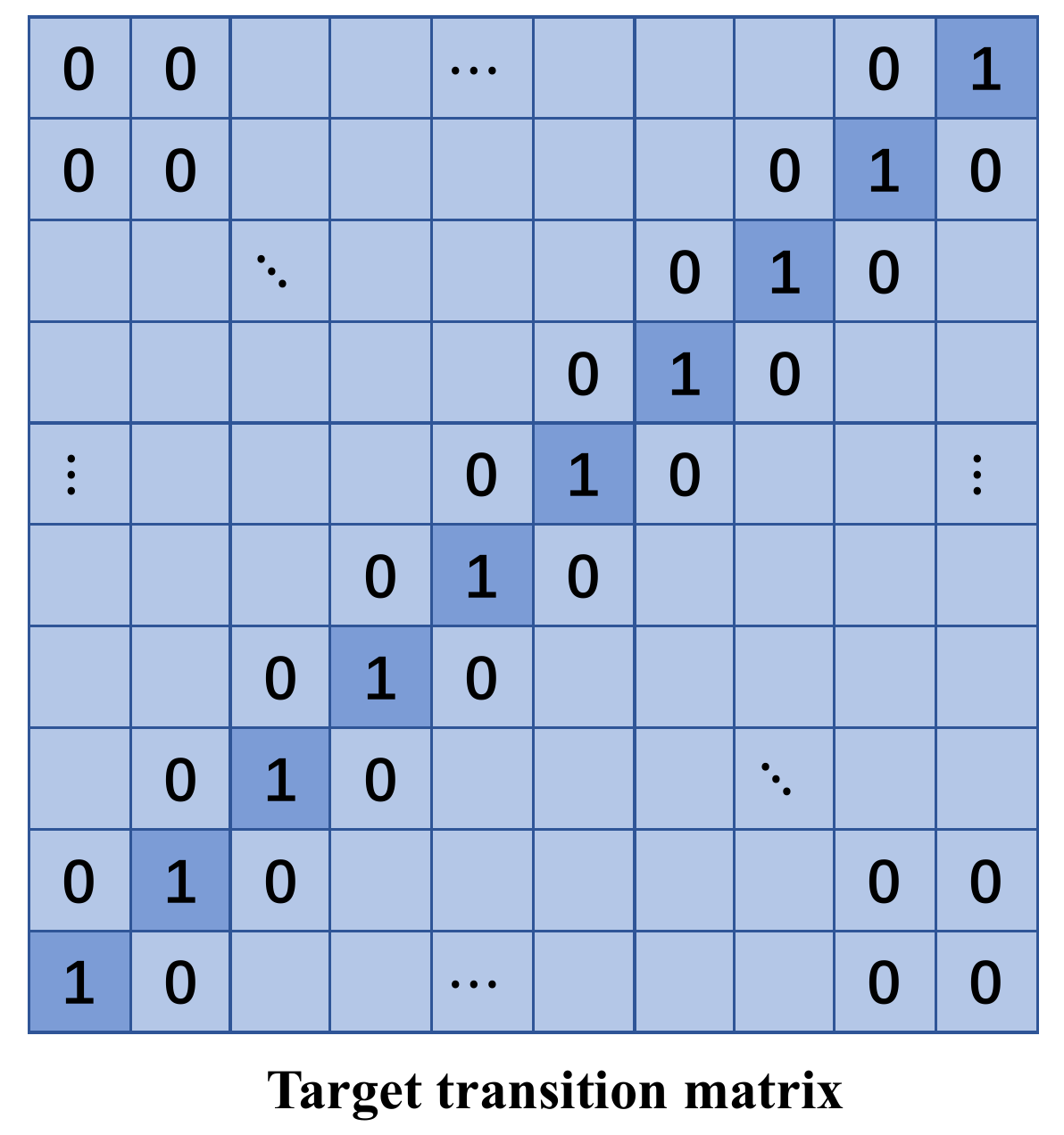}
    \caption{Target transition matrix $T^{*}$ for the target adversarial attack.}
    \label{fig5}
\end{figure*}

In scenario (ii), we design an adversarial attack to destroy the crucial transition matrix of our defense.
Since the ground-truth transition matrix is not given, we use the target attack strategy to craft adversarial examples. We choose an anti-diagonal identity matrix as an example of the target transition matrix in the target attack. The target transition matrix $T^{*}$ is shown in Fig.~\ref{fig5}. Note that there may be other attacks that can be designed, but this is beyond the scope of our work, and we would not explore further in this paper. 

\subsection{Scenario (iii): dual attack}
\label{appendix_3_3}

In scenario (iii), the optimization goal of the target attack can be designed as:
\begin{equation}
\label{eq10}
\max _{\tilde{x}} \, [-\mathcal{L}(\boldsymbol{\tilde{y}} \cdot T(\tilde{x};\omega), \boldsymbol{y^*}) + \mathcal{L}_{ce}(\boldsymbol{\tilde{y}}, \boldsymbol{y})] \text{,} \, \text{ subject to: }\left\|x-\tilde{x}\right\| \leq \epsilon \text{,}
\end{equation}
where $\boldsymbol{y^{*}}$ is the target label in the form of a vector set by the attacker. 

Note that we report the $^*\textbf{last}^*$ adversarial accuracy (i.e., the adversarial accuracy of the last epoch models obtained during training) following the discussion in \citet{carmon2019unlabeled}, instead of the $^*\textbf{best}^*$ results.

\section{The possibility of gradient masking}
\label{appendix_4}
We consider five behaviors listed in \citet{athalye2018obfuscated} to identify the gradient masking.
\textbf{(i) One-step attacks do \textit{not} perform better than iterative attacks.} The accuracy against PGD-1 is 76.81\% (vs 44.83\% against PGD-40). \textbf{(ii) Black-box attacks are \textbf{not} better (on attack success rate) than white-box attacks.} We use ResNet-18 with standard AT as the target model to craft adversarial data. The accuracy against PGD-40/AA is 70.13\%/67.30\% (vs 44.83\%/39.43\% in white-box setting). \textbf{(iii) Unbounded attacks \textbf{reach} 100\% success.} The accuracy against PGD-40 with $\epsilon=255/255$ is 0\%. \textbf{(iv) Random sampling does \textit{not} find adversarial examples.} For samples that is not successfully attacked by PGD, we randomly sample $10^5$ points within the $\epsilon$-ball, and do not find adversarial data. \textbf{(v) Increasing distortion bound \textit{increases} success.} The accuracy against PGD-40 with increasing $\epsilon$ (8/255, 16/255, 32/255 and 64/255) is 44.83\%, 26.59\%, 14.66\% and 8.12\%. These results show that our method does not use gradient masking.

\section{The influence of more model parameters}
\label{appendix_5}
We think that the improvement of the adversarial robustness has little relationship with the fact that the transition network introduces more model parameters. The transition network is only used to learn the transition matrix, and it does not directly learn the logit output for the instance to predict the class label. The main reason why our method can improve the robustness is that we explicitly model adversarial noise in the form of the transition matrix. We can use this transition matrix to infer the natural label. In addition, considering that the model parameters of our method are indeed larger than those of AT, we conduct the following experiment to compare AT and our method in the case of similar number of model parameters.

Considering that both the target model and the transition network in our method are mainly based on ResNet-18 on \textit{CIFAR-10}, we use two parallel ResNet-18 as the target model to classify the input. We use the average of their logit outputs as the final logit output. We use the training mechanism of AT to train the new target model. The target model for our MAN-based defense method is still ResNet-18. In this way, The number of model parameters of our method is similar to that of AT. As shown in Tab.~\ref{appendix_tab5}, the results show that our method still has higher adversarial accuracy. This demonstrates that the improvement of the adversarial robustness is not due to the increase of model parameters.

\begin{table*}[hbtp]
\caption{Adversarial accuracy (percentage) of defense methods against white-box adaptive attacks on \textit{CIFAR-10}. The defense models have a similar number of model parameters.}
\label{appendix_tab5}
\renewcommand\tabcolsep{4.0 pt}
\renewcommand\arraystretch{1.2}
\begin{center}
\begin{tabular}{l|cccccc}
\hline
Defense & None & PGD-40 & AA & FWA-40 & CW$_2$ & DDN \\ \hline
AT & \textbf{83.53$\pm$0.61} & 42.40$\pm$0.59& 39.09$\pm$0.60& 15.46$\pm$0.47 &0.00$\pm$0.00 & 0.08$\pm$0.01  \\
MAN & 82.72$\pm$0.53 & \textbf{44.83$\pm$0.47} & \textbf{39.43$\pm$0.73} & \textbf{29.53$\pm$0.47} &\textbf{43.17$\pm$0.68} & \textbf{10.63$\pm$0.49} \\ \hline
\end{tabular}
\vskip -0.1in
\end{center}
\end{table*}

\end{document}